\documentclass{article}
\usepackage{spconf}

\usepackage[utf8]{inputenc}
\usepackage{url}
\usepackage{graphicx}
\usepackage{authblk}
\usepackage{dirtytalk}
\usepackage{amsfonts}
\usepackage[ruled, lined]{algorithm2e}
\usepackage[square, numbers]{natbib}

\usepackage{hyperref}

\title{This changes to that : Combining causal and non-causal explanations to generate disease progression in capsule endoscopy}
\name{Anuja Vats$^{\star}$\thanks{Thanks to Research Council of Norway for funding (Project no: 300031). $^\diamond$authors contribute equally.} \quad Ahmed Mohammed$^{\star\dagger\diamond}$ \quad Marius Pedersen$^{\star\diamond}$ \quad Nirmalie Wiratunga$^{\ddagger\diamond}$}
\address{$^{\star}$ Department of Computer Science, NTNU, Gjøvik, Norway 
$^{\dagger}$Sintef Digital, Oslo, Norway
$^{\ddagger}$School of Computing, Robert Gordon University, Aberdeen, Scotland}
\begin{document}
\maketitle
\begin{abstract}
Due to the unequivocal need for understanding the decision processes of deep learning networks, both modal-dependent and model-agnostic techniques have become very popular. Although both of these ideas provide transparency for automated decision making, most methodologies focus on either using the modal-gradients (model- dependent) or ignoring the model internal states and reasoning with a model's behavior/outcome (model-agnostic) to instances. In this work, we propose a unified explanation approach that given an instance combines both model-dependent and agnostic explanations to produce an explanation set. The generated explanations are not only consistent in the neighborhood of a sample but can highlight causal relationships between image content and the outcome. We use Wireless Capsule Endoscopy (WCE) domain to illustrate the effectiveness of our explanations. The saliency maps generated by our approach are comparable or better on the softmax information score. 
\end{abstract}
\section{Introduction}
\label{sec:intro}
There has been a rapid integration of deep learning based models in real-world applications, including high risk ones such as healthcare and defence owing to their unparalleled predictive performance \cite{benjamens2020state}. Such real-world deployment and usage of models accompanies with it the moral obligation to make their decision processes transparent. 
This is necessary not only for accountability of high stake decisions but also for the identification and mitigation of algorithmic or societal bias \cite{obermeyer2019dissecting,chodosh2018courts}. This has led research to continue attempts at opening the black boxes, to gain insight in decision making processes \cite{JMLR:v11:baehrens10a, sundararajan2017IG, kapishnikov2021GIG} while also considering that useful explanations could emerge through model-agnostic explainer methods (~\cite{lime,wachter2017gdpr,byrne2019counterfactuals,discern}).
Although both of these approaches are suited to explaining model predictions, dominant explanation approaches today focus on one or the other.

Factual explainers that reason with gradients \cite{smilkov2017smoothgrad, kapishnikov2021GIG, sundararajan2017IG} aim to identify regions or pixels within an image that most significantly contributed to the prediction and thereafter visualize these attribution weights in saliency maps~\cite{JMLR:v11:baehrens10a, sundararajan2017IG, smilkov2017smoothgrad}. 
For example, given an endoscopic image with an ulcer, a saliency map would highlight the ulcer region in response to a question such as \say{Why did you make that decision?}.
However, although popular saliency methods~\cite{smilkov2017smoothgrad, sundararajan2017IG, kapishnikov2021GIG} are fairly easy to implement they also have limitations. 

One limitation of gradient-based saliency methods is the use of a baseline image and the sensitivity to the choice of that baseline~\cite{kindermans2019reliability}. 
Here a baseline helps contrast the query scenario from a \say{baseline} scenario and typically marks an absence against which the \say{presence} can be measured, e.g a black image. Since attribution maps are then accumulated over a classical linear path from baseline to the query; the choice of baseline is crucial to the success of the explanation. 
A second limitation is that the pixel perturbations done to arrive from a baseline to query are typically blind to image content. 
We argue against such pixel perturbations to create images between the baseline and query image as well as the baseline itself. Because not only are the images in between not natural but are also prone to abnormal gradient behaviours from irrelevant pixels as identified in \cite{kapishnikov2021GIG}. 

Consider our ulcer example from before, the perturbations with respect to clinical biomarkers relating to ulcer abnormality are more meaningful than individual pixels. 
For example take perturbations that cause \say{more or less inflammation around a suspected ulcer}, knowing that an ulcer is often accompanied with inflammation is important, as a lack of it might suggest incorrectly to the doctor that the suspected ulcer is just intestinal debris stuck to the surface. 
Such perturbations are not only more meaningful but every image resulting from them is directly interpretable. This also implies  that a more apt baseline would be one that marks absence of the biomarker (here the ulcer) and not complete absence of the signal. 
A third limitation of such methods is their single-pointwise explanation mode of operation \cite{alvarezmelis2018robustness, NEURIPS2018_sanitycecks}, whereby an explanation to a given image is made in isolation of its locality, i.e. without considering its neighborhood (i.e, how explanation changes as the input changes slightly).

Counterfactual reasoning has gained popularity \cite{byrne2019counterfactuals} as a locality-aware explainer that is model agnostic (i.e. does not need access to a network's internal mechanism (gradients, layer activations, etc). 
Often these provide causally understandable explanations which have been argued to be GDPR compliant \cite{wachter2017gdpr} and help address questions on fairness, trust and robustness \cite{pmlr-v97-goyal19a}. These explanations generate a counterfactual as an alternative scenario with a desirable outcome that counters the observed (real) outcome. 
As such they generate explanations through relationships like: 
\say{If the ulcer had not been present, this image would not be abnormal.} In other words, it pinpoints how the input must change to flip the outcome. It is clear how such explanations might seem intuitive and interesting \cite{byrne2019counterfactuals} to a doctor in our context. In fact, counterfactual thinking is very natural to how humans reason especially in response to negative outcomes in order to prevent them in the future \cite{roese1997counterfactual}. 

In vision, explaining an instance with its corresponding counterfactual \cite{pmlr-v97-goyal19a,alipour2022explaining} has become common for highlighting changes that would most easily flip the prediction. 
In \cite{pmlr-v97-goyal19a}, authors perform minimal edits by swapping regions of a query image from a distractor image till a decision flip occurs. 
However, the choice of a suitable distractor image is crucial for quick convergence but this choice can be unintuitive for some domains such as the medical domain or when little information is available for the dataset. Further, the image resulting from such edits can be unnatural looking at times and therefore lack explainability. For such explanations to be efficient, the changes applied to the image for a different prediction must be minimal and human interpretable \cite{wachter2017gdpr}. Alipour et al. \cite{alipour2022explaining} use the latent space of a pretrained styleGAN for retrieving counterfactual latent codes and is similar to our approach in idea but differs in implementation (their method produces causal explanations only unlike ours, while also employing pretrained attribute detectors in latent space that are largely unavailable for medical domains.)  
Recently, semi-factuals have been argued to offer advantages similar to counterfactuals \cite{Kenny_Keane_2021}. As opposed to counterfactuals that propose explanations as 'If only' clause, semi-factuals propose explanation of type 'even if' i.e. what changes to the situation would still lead to the same outcome. In our earlier example, a semi-factual image might illustrate the inflammatory changes that occur right before an ulcer starts forming, as this point the doctor will still identify the image as abnormal.

Despite advantages, one of the biggest challenges in using counterfactual and semi-factual explanations (together referred to as contrastive explanations) lies in generating instances that not only expose realistic and progressive visual changes smoothly (as to be directly understandable), but also ensuring progression alignment with the expected class prediction behaviour (congruous change in softmax score) \cite{alipour2022explaining}. 
Addressing this need for aligned progression both in the image and classifier space is precisely the problem we propose to solve in this work. 
We argue that in favor of human interpretability and algorithmic transparency, explanations that support both the aformentioned modes (causal and non-causal) are better than either one. We demonstrate the effectiveness of our explanations in the domain of WCE with focus on Ulcerative Colitis (UC). 
We use the UC biomarkers used by experts in diagnosis such as inflammation and ulcerations as progression attributes to manage counterfactual explanations. 
The main contributions are:
\begin{itemize}
\itemsep0em 
    \item a unified framework that generates both causal and non-causal explanations for each decision; 
    \item a method to control progression along a specific UC biomarker such that the counterfactual relationships inferred are causal as opposed to being adhoc; and
    \item a formal algorithm to generate saliency maps that are comparable to (or better) than others on the Softmax Information Curve (SIC) metrics \ref{sec:results}.
\end{itemize}



\begin{figure*}[!thbp]
\center
\includegraphics[]{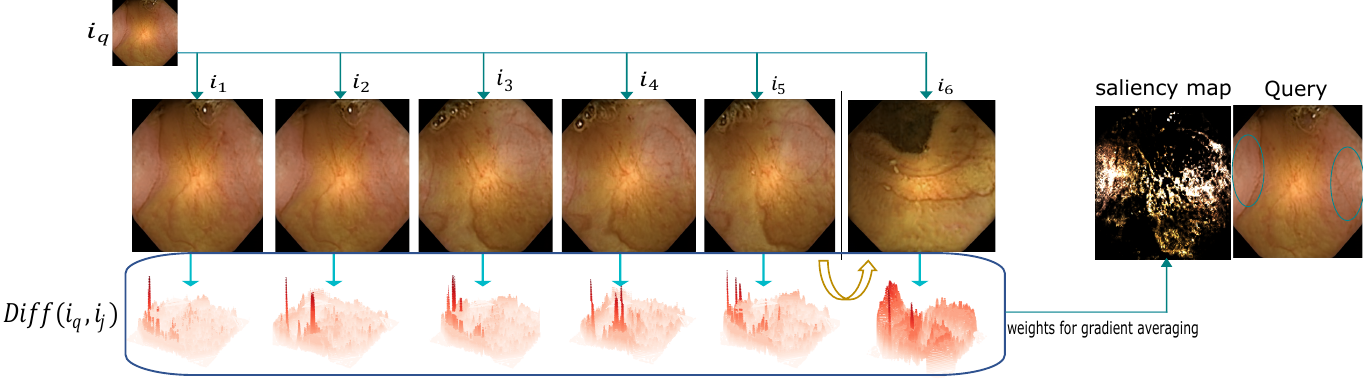}
\caption{Figure shows $i_a$ and the corresponding directional derivatives. The derivatives expose the semantic similarity between the query and it's neighbors. We use this similarity to weigh in the contribution of each neighbor towards the saliency map.}
\label{diff_map}
\end{figure*}
\section{Methodology}
\begin{figure}[!bh]
\center
\includegraphics[width=3.5in]{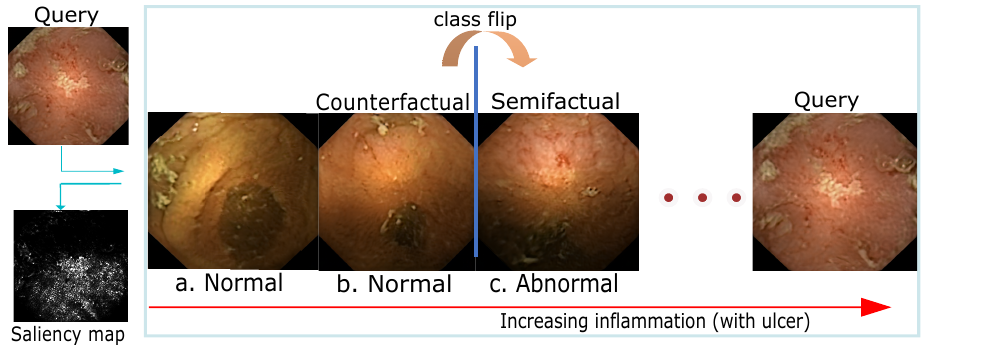}
\caption{The approach explains a query image along the ulcer attribute path together with a semifactual and counterfactual along the same path. Here the query exhibits an abnormality with inflammation. \textbf{Even} with inflammation reduced down to as in (c) the prediction would still be abnormal (semifactual). However, \textbf{if only} the visual signs change from (c) to as in (b), the prediction would be normal (counterfactual).}
\label{intro_im}
\end{figure}

Given an attribute of choice (e.g., a UC biomarker like inflammation, vascular pattern etc.) and the query image \(i^q_a\), the goal is to retrieve two instances that are closest to the decision boundary as semifactual (on the same side) and counterfactual (on the other side), while preserving visual interpretability along a path of images directed by the attribute (Figure~\ref{intro_im}). 
Regions of importance is highlighted by a saliency map (can be generated for each image on the path, including the query).

Given a classifier \(\mathcal{C}\) that outputs label \( y\in\{0,1\}\) through a prediction function \(f:\mathbb{R}^n \rightarrow [0,1]\) for an image  \(x_i \in \mathbb{R}^{512*512}\), 
an explanation set is produced, \(\mathcal{X} = \{i_{sm}, i_{cf}, i_{sf}\}\), along attribute $a$. Here \(i_{sm}\) is the saliency map, 
\(i_{cf}\) is the nearest counterfactual and \(i_{sf}\) the semifactual along \(a\). 
We use this to generate an explanation: \say{ image \(i^q_a\) is abnormal with probability \(p\) due to signs/regions highlighted by the saliency map \(i_{sm}\). The least amount of abnormality required for the prediction to be abnormal is seen in \(i_{sf}\) (semifactual). However, if the abnormal signs change to as in \(i_{cf}\) (counterfactual) the image would no longer be classified as abnormal}. 
Importantly the changes along the single attribute, $a$, is also directly visually interpretable by a user (e.g., a doctor).

\textbf{ Attribute discovery in latent space:} 
We use StyleGAN2 \cite{Karras2019stylegan2} and train it on WCE images (discussed in Dataset and Training details sec). StyleGAN2 uses a mapping network between a latent variable and the network generator, \(G\), which transforms the latent variable to an intermediate d-dimensional space, \(W\), of latent vectors, \(w\in\mathbb{R}^d\), where style attributes are known to be more amenable to control. 
We use SeFA \cite{shen2021closed} for the unsupervised discovery of attributes in the intermediate \(W\) space. In the natural image domain, pretrained attribute detectors can be utilized for labeling these attributes however for our case of pathological and anatomical variations of the colon such attribute detectors are not available a priori. 
We perform clustering on images using TSNE \cite{vanDerMaaten2008tsne} for isolating attributes relevant to pathological changes. This is done by planting seed images before clustering that had been identified by a doctor as good representatives of UC pathological changes. Upon clustering we sampled the attributes closest to seed images and have used these as explanation attributes.

\textbf{Generating the explanation set $\mathcal{X}$:}
\begin{figure}[htbp]
\center
\includegraphics[width=3.1in,height=0.9in]{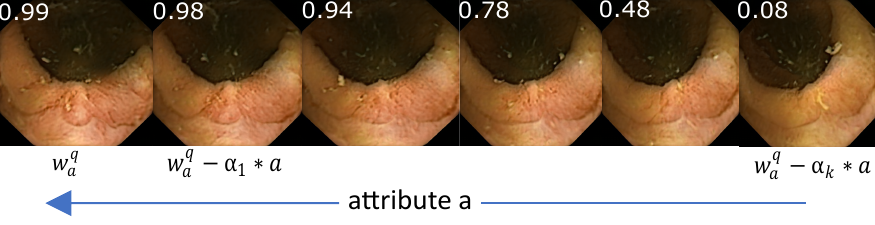}
\caption{Images in $i_a$ along attribute \(a\). Top left corner shows softmax score. Notice how apart from effected region (for attribute a), other regions in the image undergo only minimal changes. As a result, the generated explanations are consistent in the locality of a query.}
\label{example}
\end{figure}
Once relevant attributes are identified, to explain a query \(i^q_a\) with latent \(w_a^q \in \mathcal{R}^{d}\) such that \(i^q_a = G(w_a^q)\) along attribute \(a\), a set of $k$ local images $i_a$ is created,  $i_a=\{i^1_a, i^2_a ... i^k_a\}$ from latents \(w_a = \{w_a^1, w_a^2, ... w_a^k \}\) where \(w_a^j = w_a^q - \alpha_j * a\) and \(\alpha_j\) varies linearly in [A, B] and \(a \in \mathbb{R}^{512}\) is the aforementioned attribute vector. In Figure.\ref{example}, attribute \(a\) corresponds to (reddish) inflammatory regions and set \(i_a\) can be understood as images with decrease in severity of such inflammation as \(\alpha\) progresses from \(A=0\) to \(B=30\). \(i_{cf}, i_{sf}\) in \(\mathcal{X}\) are retrieved based on the classifier output for \(i_a\) such that \( i_{cf} = argmax(\sigma(\mathcal{C}(i_a^j))) \hspace{0.25em}\forall \hspace{0.25em} {\sigma(\mathcal{C}(i_a^j))<0.5}\) and \(i_{sf} = argmin(\sigma(\mathcal{C}(i_a^j))) \hspace{0.25em}\forall\hspace{0.25em} {\sigma(\mathcal{C}(i_a^j))>0.5}\) where \(\sigma\) is the softmax function. For the saliency map, to avoid the spuriousness observed in previous literature, we use the latent space to curate a neighborhood such that every image in the neighborhood of a query varies only along the chosen attribute. In other words, the pixel changes that occur in this neighborhood are neither uniform nor content blind \cite{kapishnikov2021GIG, sundararajan2017IG}, but targeted towards those pixels that most strongly affect the attribute/biomarker. We use directional derivatives in $i_a$ along attribute $a$ for identifying these regions and weight them based on semantic similarity with $i^q_a$ to generate the saliency map. The directional derivative $Diff(i_a^q, i_a^j)$  between the query and $i_a^j = G(w_a^j) \forall \{w_a^j\}_{j=1}^k$ is given by:
\begin{equation}
\label{eq:1}
 \mathit{Diff}(i_a^q, i_a^j) = \left|\frac{ G(w_a^q) - G(w_a^j)}{1} \right|
\end{equation}
The directional derivatives $\mathit{Diff}(i_a^q, i_a^j)$ over $i^q_a$ and $i_a$ exposes pixels with consistent change in the direction of increasing/decreasing attribute (see Figure. \ref{diff_map}), in other words it is a measure of semantic similarity to the image being explained. We use these derivatives to measure the contribution of each image in $i_a$. A formal algorithm is described in algorithm \ref{alg:smap}.

\begin{algorithm}[htbp]
\caption{Saliency map generation}
\label{alg:smap}
\scriptsize
	https://www.overleaf.com/project/632c20113b1a4fb9b68a2cfb\KwIn{Classifier C, query $i^q_a$; $i_a=\{i^1_a, i^2_a ... i^k_a, i^q_a\}$;}
	\KwOut{$i_{sm}$}
	\ForEach{$i^q_a$}{
	    predict output class probabilities for $i_a$\\
		$output \gets C(i_a)$ \\
		backpropagate and collect gradients wrto $i_a$\\
		$[grad^1_a, grad^2_a ... grad^q_a] \gets i_a.grad()$\\
		directional derivatives along attribute a\\
		\ForEach{$i \in (i_a \backslash \{i^q_a\})$}{
		$Diff(i_q, i)  \gets \left|\frac{ G(w_a^q) - G(w_a^i)}{1} \right|$
		}
		$S(i_q, a) \gets \frac{ \sum_{j=1}^{k} grad_a^j \cdot Diff(i^q, j)}{k}$\\
		$i_{sm}  \gets meanThresholding(S(i_q, a))$\\
	}
\end{algorithm}

\textbf{ Dataset and Training Details:}\\
\label{dataset}
The dataset consists of approximately 200k unlabeled WCE images. The majority of images come from  WCE examinations of 10 patients with varying UC activity, as well as other pathologies with PillCam Colon 2 Capsule, Medtronic. The images are 576x576 in resolution with varying degree of bowel cleanliness. In addition to this we use PS-DeVCEM dataset\cite{AnujaNature} with 80k images of the same capsule modality. Remaining images come from the OSF-Kvasir Dataset \cite{smedsrud2021kvasir} with 3478 images from seven classes taken with the capsule modality Olympus EC-S10. We use StyleGAN2 without progressive growing and work exclusively on the original intermediate latent space \(W\) and not the extended space \(W^+\). The model was trained on TwinTitan RTX for 30 days.\footnote{Github : https://github.com/anuja13/ContrastiveExplanations}
\section{Results}
\label{sec:results}
\begin{figure}[!thbp]
\center
\includegraphics[width=2.3in,height=2.1in]{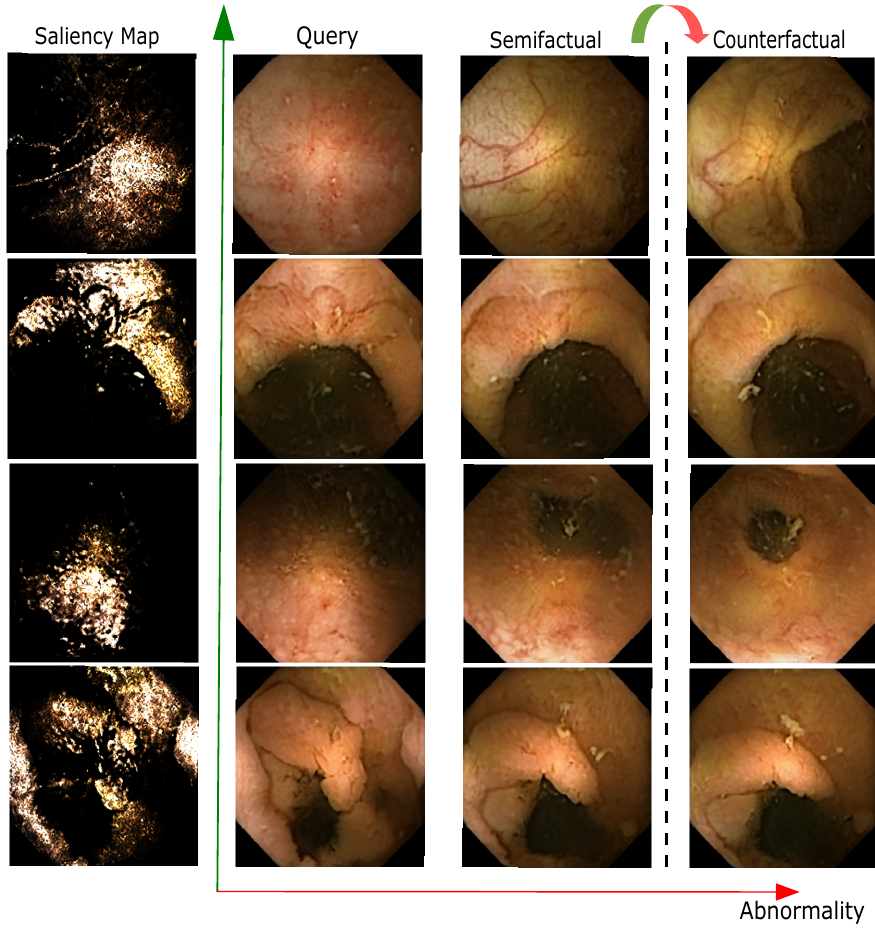}
\vspace*{-5mm}
\caption{Figure shows $\mathcal{X}$ generated with this approach on different query images (column 3). Best viewed in color.}
\label{Results}
\end{figure}

\textbf{Qualitative Comparison:} GuidedIG \cite{kapishnikov2021GIG} produces noisy saliency maps (as only pixels with low partial derivatives are moved towards their original intensity at each step to avoid high gradient regions and thus abnormal behavior), but if the pixels affecting the decision are not localized but spread globally across the image, (as in WCE), the resulting saliency map can appear to be noisier. Similarly, while Smoothgrad\cite{smilkov2017smoothgrad} captures the right regions, the saliency maps are overall noisy. Integrated Gradients \cite{sundararajan2017IG} correlates very closely with our maps. Figure.\ref{Results} shows $\mathcal{X}$ for various query images.\\
\begin{figure}[htbp]
\center
\includegraphics[width=3in,height=2in]{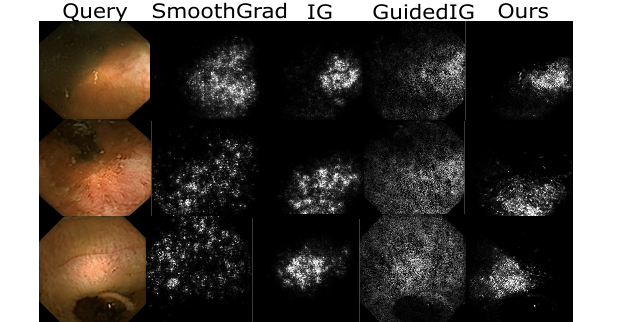}
\caption{Qualitative comparison of saliency maps between our approach and other approaches. Integrated Gradients (IG) \cite{sundararajan2017IG}, Guided integrated gradients \cite{kapishnikov2021GIG}, SmoothGrad \cite{smilkov2017smoothgrad}}
\label{smaps}
\end{figure}

\begin{figure}[htbp]
\center
\includegraphics[width=3.3in,height=2.1in]{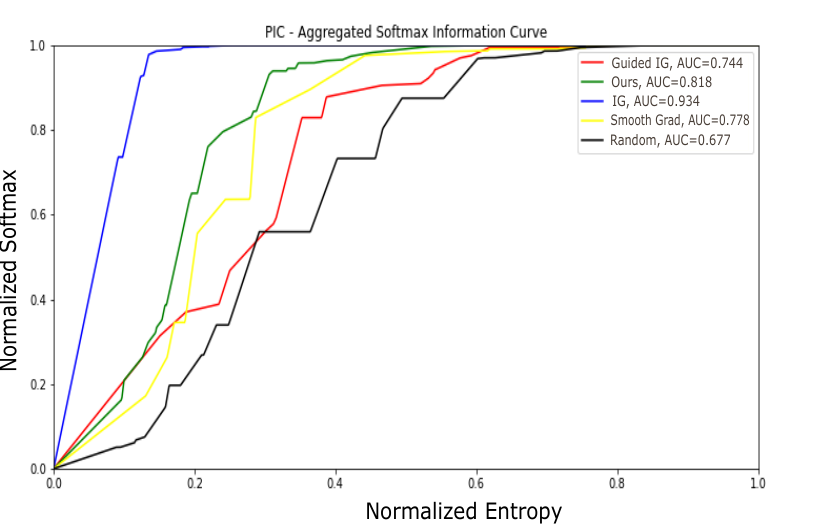}
\caption{Quantitative comparison of saliency maps : Median Softmax Information curves}
\label{SIC}
\end{figure}

\textbf{Quantitative Comparison:} We use Softmax Information Curve (SIC AUC) \cite{kapishnikov2019xrai} for quantitative comparison. SIC AUC measures the softmax score of a model against salient regions indicated by the saliency map. Figure \ref{SIC} shows the SIC AUC for different approaches averaged over 50 images. Integrated gradients achieve the best score followed by our approach. We suspect this to be due to the SIC score's preference for smallest regions of effect (as in IG) instead of identifying all contributing regions (as in ours).

\section{Conclusion}
In this work, we propose a framework for generating causal as well as non-causal explanations for any image classifier. Our model is network agnostic and supports not only visual insight into model decisions, but offers end users the opportunity to visualize alternate scenarios relevant to the current situation, for example in prognosis of UC as shown in this work. To the best of our knowledge, this is one of the first works to propose a single framework for generating both causal as well as non-causal explanation for deep learning based models. 
\pagebreak

\bibliographystyle{abbrvnat}
\bibliography{references}

\end{document}